\documentclass[letterpaper, 10 pt, conference]{ieeeconf}
\IEEEoverridecommandlockouts 

\usepackage{microtype}

\usepackage[pdftex,pdfauthor={Michael Everett, Justin Miller, and Jonathan P.\ 	How},pdftitle={Planning Beyond The Sensing Horizon Using a Learned Context}]{hyperref}
\hypersetup{colorlinks,linkcolor={green!10!black},citecolor={green!10!black},urlcolor={blue!80!black}}
\makeatletter \let\NAT@parse\undefined \makeatother
\usepackage[sort,compress]{cite}
\usepackage{graphicx} 
\usepackage{amsfonts}
\usepackage{amsmath,soul}
\usepackage{color}
\usepackage[sort,compress]{cite}
\usepackage[font=footnotesize]{subcaption}
\usepackage{balance}
\usepackage[font=footnotesize]{caption}
\usepackage[linesnumbered,ruled,vlined]{algorithm2e}
\usepackage{multirow}

\DeclareMathOperator*{\argmax}{\arg\!\max}
\usepackage{tabulary}
\newcolumntype{K}[1]{>{\centering\arraybackslash}p{#1}}

\title{\LARGE \bf Planning Beyond The Sensing Horizon Using a Learned Context}

\author{Michael Everett$^\dag$, Justin Miller$^\ddag$ and Jonathan P.\ How$^\dag$
  \thanks{$^\dag$Aerospace Controls Laboratory,
    Massachusetts Institute of Technology, 77 Massachusetts Ave.,
    Cambridge, MA, USA. {\tt\footnotesize \{mfe, jhow\}@mit.edu}}%
   \thanks{$^\ddag$Robotics and Intelligent Vehicles, Ford Motor Company, Dearborn, MI, USA. {\tt\footnotesize jmill597@ford.com }}
   \thanks{\textbf{Open-Source Software:} \url{https://github.com/mit-acl/dc2g}}
}

\usepackage[svgnames]{xcolor} \definecolor{DarkGreen}{rgb}{0,0.5,0}
\definecolor{DarkRed}{rgb}{0.75,0,0}

\usepackage[authormarkuptext=name,addedmarkup=bf,authormarkupposition=left]{changes}
\definechangesauthor[name={J.~H.}, color={blue}]{jh}
\definechangesauthor[name={M.~E.}, color={red}]{me}

\usepackage[super]{nth}
\usepackage{tikz,mathtools}
\usepackage[capitalize]{cleveref}
\crefformat{equation}{(#2#1#3)}
\Crefformat{equation}{Equation~(#2#1#3)}
\Crefname{equation}{Equation}{Equations}

\usetikzlibrary{shapes,positioning,automata,arrows,fit,backgrounds,calc}
\tikzstyle{block} = [draw, fill=blue!20, rectangle,minimum height=1em,
minimum width=2em] \tikzstyle{sum} = [draw, fill=blue!20, circle, node
distance=1cm] \tikzstyle{input} = [coordinate] \tikzstyle{output} =
[coordinate] \tikzstyle{pinstyle} = [pin edge={to-,thin,black}]
\usetikzlibrary{trees} \usetikzlibrary{decorations.pathmorphing}
\usetikzlibrary{decorations.markings}
\definecolor{darkgreen}{rgb}{0,0.5,0}
\definecolor{darkred}{rgb}{220,20,60}

\makeatletter
\renewcommand\paragraph{\@startsection{subsubsection}{4}{\z@}%
{0.25ex \@plus.5ex \@minus.2ex}%
{-.15em}%
{\normalfont\normalsize\itshape}}
\makeatother

\usepackage{setspace}
\begin{document}

\maketitle
\thispagestyle{empty} \pagestyle{empty}

\begin{abstract}
Last-mile delivery systems commonly propose the use of autonomous robotic vehicles to increase scalability and efficiency.
The economic inefficiency of collecting accurate prior maps for navigation motivates the use of planning algorithms that operate in unmapped environments.
However, these algorithms typically waste time exploring regions that are unlikely to contain the delivery destination.
Context is key information about structured environments that could guide exploration toward the unknown goal location, but the abstract idea is difficult to quantify for use in a planning algorithm.
Some approaches specifically consider contextual relationships between objects, but would perform poorly in object-sparse environments like outdoors.
Recent deep learning-based approaches consider context too generally, making training/transferability difficult.
Therefore, this work proposes a novel formulation of utilizing context for planning as an image-to-image translation problem, which is shown to extract terrain context from semantic gridmaps, into a metric that an exploration-based planner can use.
The proposed framework has the benefit of training on a static dataset instead of requiring a time-consuming simulator.
Across 42 test houses with layouts from satellite images, the trained algorithm enables a robot to reach its goal 189\% faster than with a context-unaware planner, and within 63\% of the optimal path computed with a prior map.
The proposed algorithm is also implemented on a vehicle with a forward-facing camera in a high-fidelity, Unreal simulation of neighborhood houses.
\end{abstract}

\section{Introduction} \label{sec:intro}
A key topic in robotics is the use of automated robotic vehicles for last-mile delivery.
A standard approach is to visit and map delivery environments ahead of time, which enables the use of planning algorithms that guide the robot toward a specific goal coordinate in the map.
However, the economic inefficiency of collecting and maintaining maps, the privacy concerns of storing maps of people's houses, and the challenges of scalability across a city-wide delivery system are each important drawbacks of the pre-mapping approach.
This motivates the use of a planning framework that does not need a prior map.
In order to be a viable alternative framework, the time required for the robot to locate and reach its destination must remain close to that of a prior-map-based approach.

\begin{figure}[t]
	\centering
	\begin{subfigure}{0.33\columnwidth}
		\centering
		\includegraphics [trim=0 0 0 0, clip, width=\textwidth, angle = 0]{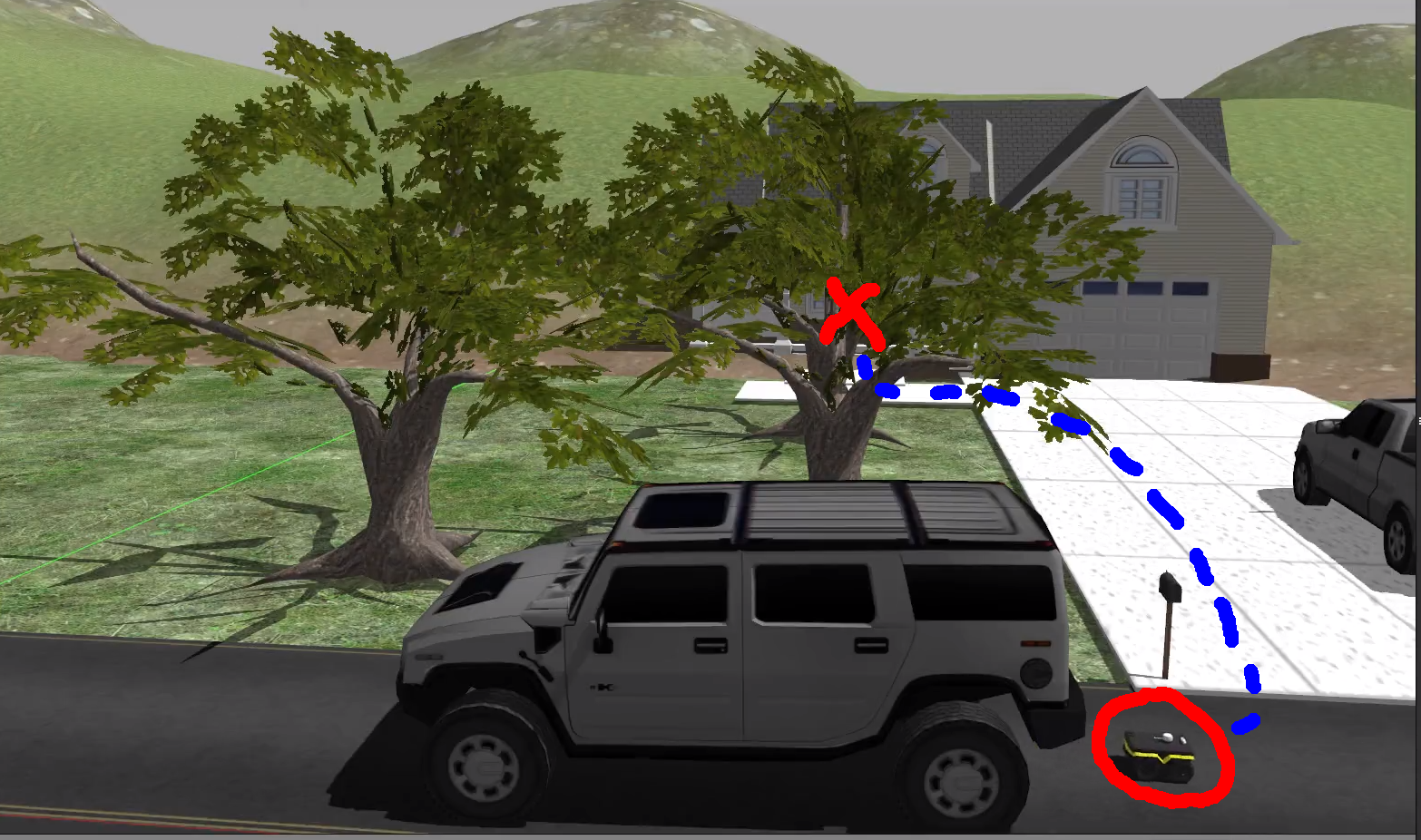}
		\caption{Oracle's view}
		\label{fig:robot_delivery_scene} 
	\end{subfigure}
	\begin{subfigure}{0.31\columnwidth}
		\centering
		\includegraphics [trim=0 0 0 0, clip, width=\textwidth, angle = 0]{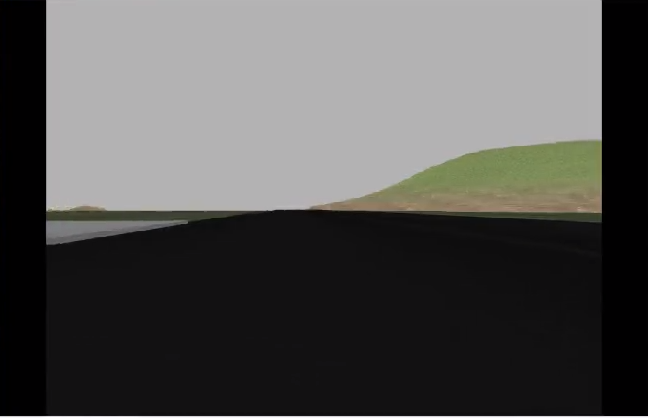}
		\caption{Robot's view}
		\label{fig:robot_delivery_camera} 
	\end{subfigure}
	\begin{subfigure}{0.31\columnwidth}
		\centering
		\includegraphics [trim=0 0 0 0, clip, width=\textwidth, angle = 0]{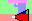}
		\caption{Semantic Map}
		\label{fig:semantic_map} 
	\end{subfigure}
	\caption{Robot delivers package to front door. If the robot has no prior map and does not know where the door is, it must quickly search for its destination. Context from the onboard camera view (b) can be extracted into a lower-dimensional semantic map (c), where the white robot can see terrain within its black FOV.}
	\label{fig:rover}
\end{figure}

Consider a robot delivering a package to a new house's front door (\cref{fig:rover}).
Many existing approaches require delivery destinations to be specified in a format useful to the robot (e.g., position coordinates, heading/range estimates, target image), but collecting this data for every destination presents the same limitations as prior mapping.
Therefore the destination should be a high-level concept, like ``go to the front door.''
Such a destination is intuitive for a human, but without actual coordinates, difficult to translate into a planning objective for a robot.
The destination will often be beyond the robot's economically-viable sensors' limited range and field of view.
Therefore, the robot must explore~\cite{yamauchi1998frontier,stachniss2005information} to find the destination; however, pure exploration is slow because time is spent exploring areas unlikely to contain the goal.
Therefore, this paper investigates the problem of efficiently planning beyond the robot's line-of-sight by utilizing \textit{context} within the local vicinity.
Existing approaches use context and background knowledge to infer a geometric understanding of the high-level goal's location, but the representation of background knowledge is either difficult to plan from~\cite{joho2011learning,samadi2012using,kollar2009utilizing,kunze2014bootstrapping,kunze2014using,lorbach2014prior,vernaza2014learning,aydemir2013active,hanheide2017robot}, or maps directly from camera image to action~\cite{brahmbhatt2017deepnav,wu2018building,blukis2018following,zhu2017icra,zhu2017visual,gordon2018iqa,gupta2017cognitive,embodiedqa}, reducing transferability to real environments.

This work proposes a solution to efficiently utilize context for planning.
Scene context is represented in a semantic map, then a learned algorithm converts context into a search heuristic that directs a planner toward promising regions in the map.
The context utilization problem (determination of promising regions to visit) is uniquely formulated as an image-to-image translation task, and solved with U-Net/GAN architectures~\cite{ronneberger2015u,pix2pix2017} recently shown to be useful for geometric context extraction~\cite{pronobis2017learning}.
By learning with semantic gridmap inputs instead of camera images, the planner proposed in this work could be more easily transferred to the real world without the need for training in a photo-realistic simulator.
Moreover, a standard local collision avoidance algorithm can operate in conjunction with this work's global planning algorithm, making the framework easily extendable to environments with dynamic obstacles.

The contributions of this work are
i) a novel formulation of utilizing context for planning as an image-to-image translation problem, which converts the abstract idea of scene context into a planning metric,
ii) an algorithm to efficiently train a cost-to-go estimator on typical, partial semantic maps, which enables a robot to learn from a static dataset instead of a time-consuming simulator,
iii) demonstration of a robot reaching its goal 189\% faster than a context-unaware algorithm in simulated environments, with layouts from a dataset of real last-mile delivery domains,
and
iv) an implementation of the algorithm on a vehicle with a forward-facing RGB-D + segmentation camera in a high-fidelity simulation.

\begin{figure*}[t]
	\centering
	\includegraphics [trim=0 300 0 0, clip, width=0.9\linewidth, angle = 0, page=2]{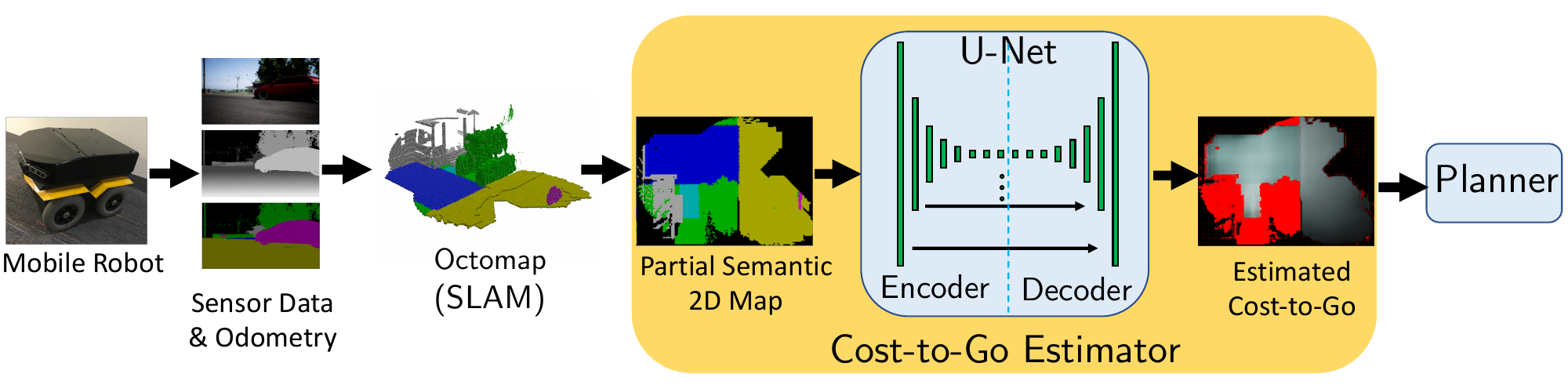}
	\caption{System architecture. To plan a path to an unknown goal position beyond the sensing horizon, a robot's sensor data is used to build a semantically-colored gridmap. The gridmap is fed into a U-Net to estimate the cost-to-go to reach the unknown goal position. The cost-to-go estimator network is trained offline with annotated satellite images. During online execution, the map produced by a mobile robot's forward-facing camera is fed into the trained network, and cost-to-go estimates inform a planner of promising regions to explore.}
	\label{fig:system_architecture}
	\vspace{-0.2in}
\end{figure*}

\section{Related Work}\label{sec:related_work}

\subsubsection{Planning \& Exploration}
Classical planning algorithms rely on knowledge of the goal coordinates (A*, RRT) and/or a prior map (PRMs, potential fields), which are both unavailable in this problem.
Receding-horizon algorithms are inefficient without an accurate heuristic at the horizon, typically computed with goal knowledge.
Rather than planning to a destination, the related field of exploration~\cite{yamauchi1998frontier,stachniss2005information} is a conservative search strategy, and pure exploration algorithms often estimate \textit{information gain} of planning options using geometric context.
However, exploration and search objectives differ, meaning the exploration robot will spend time gaining information in places that are useless for the search task.

\subsubsection{Context for Object Search}
Leveraging scene context is therefore fundamental to enable object search that outperforms pure exploration.
Many papers consider a single form of context.
Geometric context (represented in occupancy grids) is used in~\cite{6907760,10.1007/978-3-319-50115-4_34,Richter2018,bhardwaj2017learning}, but these works also assume knowledge of the goal location for planning.
Works that address true object search usually consider semantic object relationships as a form of context instead.
Decision trees and maximum entropy models can be trained on object-based relationships, like positions and attributes of common items in grocery stores~\cite{joho2011learning}.
Because object-based methods require substantial domain-specific background knowledge, some approaches automate the background data collection process by using Internet searches~\cite{samadi2012using,kollar2009utilizing}.
Object-based approaches have also noted that the spatial relationships between objects are particularly beneficial for search (e.g., keyboards are often \textit{on} desks)~\cite{kunze2014using,kunze2014bootstrapping,lorbach2014prior}, but are not well-suited to represent geometries like floorplans/terrains.
Hierarchical planners improve search performance via a human-like ability to make assumptions about object layouts~\cite{aydemir2013active,hanheide2017robot}.

To summarize, existing uses of context either focus on relationships between objects or the environment's geometry.
These approaches are too specific to represent the combination of various forms of context that are often needed to plan efficiently.

\subsubsection{Deep Learning for Object Search}
Recent works use deep learning to represent scene context.
Several approaches consider navigation toward a semantic concept (e.g., go to the kitchen) using only a forward-facing camera.
These algorithms are usually trained end-to-end (image-to-action) by supervised learning of expert trajectories~\cite{brahmbhatt2017deepnav,DBLP:journals/corr/abs-1801-02209,blukis2018following} or reinforcement learning in simulators~\cite{zhu2017icra,zhu2017visual,gordon2018iqa,gupta2017cognitive,embodiedqa}.
Training such a general input-output relationship is challenging; therefore, some works divide the learning architecture into a deep neural network for each sub-task (e.g., mapping, control, planning)~\cite{gupta2017cognitive,blukis2018following,gordon2018iqa}.

Still, the format of context in existing, deep learning-based approaches is too general.
The difficulty in learning how to extract, represent, and use context in a generic architecture leads to massive computational resource and time requirements for training.
In this work, we reduce the dimensionality (and therefore training time) of the learning problem by first leveraging existing algorithms (semantic SLAM, image segmentation) to extract and represent context from images; thus, the learning process is solely focused on context utilization.
A second limitation of systems trained on simulated camera images, such as existing deep learning-based approaches, is a lack of transferability to the real world.
Therefore, instead of learning from simulated camera images, this work's learned systems operate on semantic gridmaps which could look identical in the real world or simulation.

\subsubsection{Reinforcement Learning}
Reinforcement learning (RL) is a commonly proposed approach for this type of problem~\cite{zhu2017icra,zhu2017visual,gordon2018iqa,gupta2017cognitive,embodiedqa}, in which experiences are collected in a simulated environment.
However, in this work, the agent's actions do not affect the static environment, and the agent's observations (partial semantic maps) are easy to compute, given a map layout and the robot's position history.
This work's approach is a form of model-based learning, but learns from a static dataset instead of environment interaction.

\subsubsection{U-Nets for Context Extraction}
This work's use of U-Nets~\cite{ronneberger2015u,pix2pix2017,pix2pix-tensorflow} is motivated by experiments that show generative networks can imagine unobserved regions of occupancy gridmaps, suggesting that they can be trained to extract significant geometric context in structured environments~\cite{pronobis2017learning}.
However, the focus of that work is on models' abilities to encode context, not context utilization for planning.


\section{Approach} \label{sec:approach}
The input to this work's architecture (\cref{fig:system_architecture}) is a RGB-D camera stream with semantic mask, which is used to produce a partial, top-down, semantic gridmap.
An image-to-image translation model is trained to estimate the planning cost-to-go, given the semantic gridmap.
Then, the estimated cost-to-go is used to inform a frontier-based exploration planning framework.

\subsection{Training Data}\label{sec:training_data}

\begin{figure*}[t]
	\centering
	\includegraphics [trim=0 230 0 0, clip, width=0.8\textwidth, angle = 0, page=1]{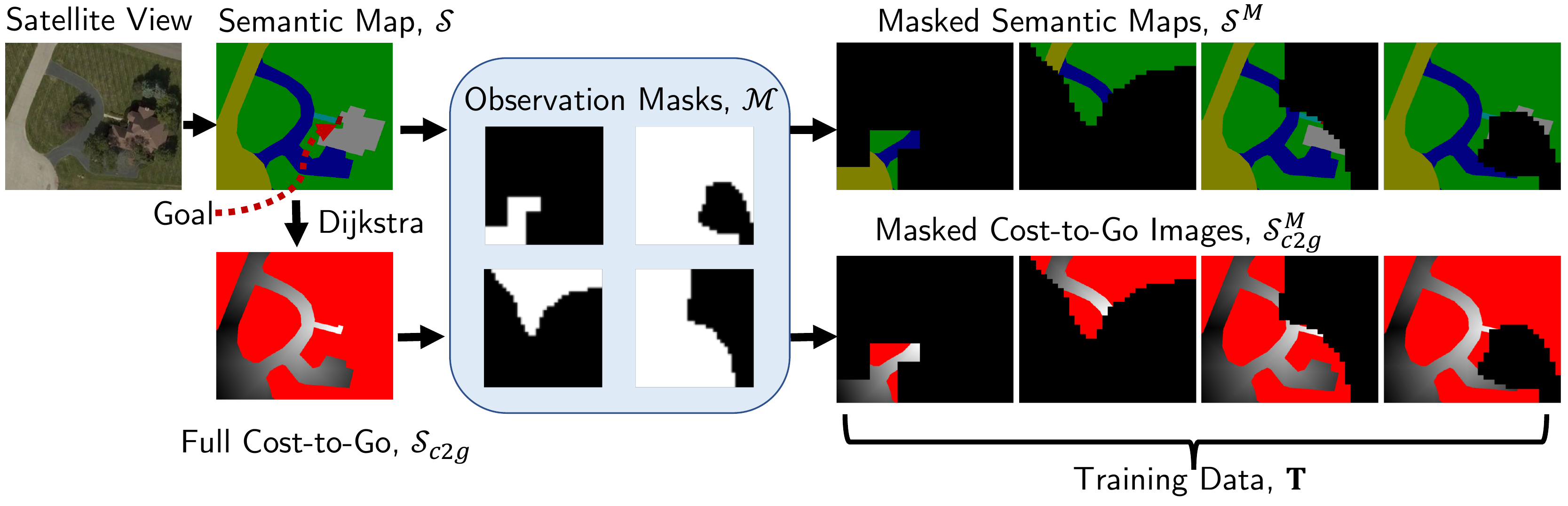}
	\caption{Training data creation. A satellite view of a house's front yard is manually converted to a semantic map (top left), with colors for different objects/terrain. Dijkstra's algorithm gives the ground truth distance from the goal to every point along drivable terrain (bottom left)~\cite{dijkstra1959note}. This cost-to-go is represented in grayscale (lighter near the goal), with red pixels assigned to untraversable regions. To simulate partial map observability, 256 observation masks (center) are applied to the full images to produce the training set.}
	\label{fig:training_data}
\end{figure*}

A typical criticism of learning-based systems is the challenge and cost of acquiring useful training data.
Fortunately, domain-specific data already exists for many exploration environments (e.g., satellite images of houses, floor plans for indoor tasks); however, the format differs from data available on a robot with a forward-facing camera.
This work uses a mutually compatible representation of gridmaps segmented by terrain class.

A set of satellite images of houses from Bing Maps~\cite{bing_maps} was manually annotated with terrain labels.
The dataset contains 77 houses (31 train, 4 validation, 42 test) from 4 neighborhoods (3 suburban, 1 urban); one neighborhood is purely for training, one is split between train/val/test, and the other 2 neighborhoods (including urban) are purely for testing.
Each house yields many training pairs, due to the various partial-observability masks applied, so there are 7936 train, 320 validation, and 615 test pairs (explained below).
An example semantic map of a suburban front yard is shown in the top left corner of~\cref{fig:training_data}.

\cref{alg:training} describes the process of automatically generating training data from a set of semantic maps, $\mathbf S$.
A semantic map, $\mathcal S \in \mathbf S$, is first separated into traversable (roads, driveways, etc.) and non-traversable (grass, house) regions, represented as a binary array, $\mathcal S_{tr}$~(\cref{alg:training:traversable}).
Then, the shortest path length between each traversable point in $\mathcal S_{tr}$ and the goal is computed with Dijkstra's algorithm~\cite{dijkstra1959note}~(\cref{alg:training:c2g}).
The result, $\mathcal S_{c2g}$, is stored in grayscale (\cref{fig:training_data} bottom left: darker is further from goal).
Non-traversable regions are red in $\mathcal S_{c2g}$.

This work allows the robot to start with no (or partial) knowledge of a particular environment's semantic map; it observes (uncovers) new areas of the map as it explores the environment.
To approximate the partial maps that the robot will have at each planning step, full training maps undergo various masks to occlude certain regions.
For some binary observation mask, $\mathcal M$: the full map, $\mathcal S$, and full cost-to-go, $\mathcal S_{c2g}$, are masked with element-wise multiplication as $\mathcal S^M = \mathcal S \circ \mathcal M$ and $\mathcal S^M_{c2g} = \mathcal S_{c2g} \circ \mathcal M$ (\cref{alg:training:mask_obs,alg:training:mask_c2g}).
The $(\textrm{input, target output})$ pairs used to train the image-to-image translator are the set of $(\mathcal S^M, \mathcal S^M_{c2g})$~(\cref{alg:training:input_output_pairs}).

\begin{algorithm}[t]
	\textbf{Input:} semantic maps, $\mathbf{S}$; observability masks, $\mathbf{M}$\\
	\textbf{Output:} training image pairs, $\mathbf{T}$ \\
	\
	\ForEach{$\mathcal S \in \mathbf S$}{
		$\mathcal S_{tr} \leftarrow$ Find traversable regions in $\mathcal S$\label{alg:training:traversable}\\
		$\mathcal S_{c2g} \leftarrow$ Compute cost-to-go to goal of all pts in $\mathcal S_{tr}$\label{alg:training:c2g}\\
		\ForEach{$\mathcal M \in \mathbf M$}{
			$\mathcal S_{c2g}^{M} \leftarrow$ Apply observation mask $\mathcal M$ to $\mathcal S_{c2g}$\label{alg:training:mask_c2g}\\
			$\mathcal S^{M} \leftarrow$ Apply observation mask $\mathcal M$ to $\mathcal S$\label{alg:training:mask_obs}\\
			$\mathbf T \leftarrow \mathcal \{(S^{M}, S_{c2g}^{M})\} \cup \mathbf T$\label{alg:training:input_output_pairs}
		}
	}
	\caption{Automated creation of NN training data}
	\label{alg:training}
\end{algorithm}

\subsection{Offline Training: Image-to-Image Translation Model}

The motivation for using image-to-image translation is that i) a robot's sensor data history can be compressed into an image (semantic gridmap), and ii) an estimate of cost-to-go at every point in the map (thus, an image) enables efficient use of receding-horizon planning algorithms, given only a high-level goal (``front door'').
Although the task of learning to predict just the goal location is easier, a cost-to-go estimate implicitly estimates the goal location and then provides substantially more information about how to best reach it.

The image-to-image translator used in this work is based on~\cite{pix2pix2017,pix2pix-tensorflow}.
The translator is a standard encoder-decoder network with skip connections between corresponding encoder and decoder layers (\mbox{``U-Net''})~\cite{ronneberger2015u}.
The objective is to supply a 256x256 RGB image (semantic map, $\mathcal S$) as input to the encoder, and for the final layer of the decoder to output a 256x256 RGB image (estimated cost-to-go map, $\mathcal{\hat{S}}_{c2g}$).
Three U-Net training approaches are compared: pixel-wise $L_1$ loss, GAN, and a weighted sum of those two losses, as in~\cite{pix2pix2017}.

A partial map could be associated with multiple plausible cost-to-gos, depending on the unobserved parts of the map.
Without explicitly modeling the distribution of cost-to-gos conditioned on a partial semantic map, $P(~\mathcal S_{c2g}^{M}~\mid~\mathcal{S}^M)$, the network training objective and distribution of training images are designed so the final decoder layer outputs a most likely cost-to-go-map, $\mathcal{\bar{S}}_{c2g}^{M}$, where
\begin{equation}
\mathcal{\bar{S}}_{c2g}^{M} = \argmax_{\mathcal S_{c2g}^{M}\in\mathbb{R}^{256\times256\times3}} P(\mathcal S_{c2g}^{M}\mid\mathcal S^M).
\end{equation}


 
\subsection{Online Mapping: Semantic SLAM}


To use the trained network in an online sense-plan-act cycle, the mapping system must produce top-down, semantic maps from the RGB-D camera images and semantic labels available on a robot with a forward-facing depth camera and an image segmentation algorithm~\cite{he2018mask}.
Using~\cite{semantic_slam}, the semantic mask image is projected into the world frame using the depth image and camera parameters to produce a pointcloud, colored by the semantic class of the point in 3-space.
Each pointcloud is added to an octree representation, which can be converted to a octomap (3D occupancy grid) on demand, where each voxel is colored by the semantic class of the point in 3-space.
Because this work's experiments are 2D, we project the octree down to a 2D semantically-colored gridmap, which is the input to the cost-to-go estimator.

\subsection{Online Planning: Deep Cost-to-Go}

This work's planner is based on the idea of frontier exploration~\cite{yamauchi1998frontier}, where a frontier is defined as a cell in the map that is observed and traversable, but whose neighbor has not yet been observed.
Given a set of frontier cells, the key challenge is in choosing $\textit{which}$ frontier cell to explore next.
Existing algorithms often use geometry (e.g., frontier proximity, expected information gain based on frontier size/layout); we instead use context to select frontier cells that are expected to lead toward the destination.

\begin{algorithm}[t]
	\textbf{Input:} current partial semantic map $\mathcal{S}$, pose $(p_x, p_y, \theta)$\\
	\textbf{Output:} action $\mathbf{u}_t$ \\
	$\mathcal S_{tr} \leftarrow$ Find traversable cells in $\mathcal S$~\label{dc2g:traversable}\\
	$\mathcal S_{r} \leftarrow$ Find reachable cells in $\mathcal S_{tr}~\label{dc2g:reachable}$ from $(p_x, p_y)$ w/ BFS\\
	\If{goal $\not\in \mathcal S_{r}$}{~\label{dc2g:if_goal_reachable}
		$\mathcal F \leftarrow$ Find frontier cells in $\mathcal S_r$~\label{dc2g:frontier}\\
		$\mathcal F^e \leftarrow$ Find cells in $\mathcal S_r$ where $f \in \mathcal F$ in view~\label{dc2g:frontier_expanding}\\
		$\mathcal F_r^e \leftarrow$ $\mathcal S_r \cap \mathcal F^e$: reachable, frontier-expanding cells~\label{dc2g:frontier_reachable_expanding}\\
		$\mathcal{\hat{S}}_{c2g} \leftarrow$ Query generator network with input $\mathcal S$~\label{dc2g:query}\\
		$\mathcal C \leftarrow$ Filter and resize $\mathcal{\hat{S}}_{c2g}$~\label{dc2g:filter_and_resize}\\
		$(f_x, f_y) \leftarrow$ $\argmax_{f\in\mathcal F_r^e} \mathcal C$~\label{dc2g:min_cost}\\
		$\mathbf{u}_{t:\infty} \leftarrow$ Backtrack from $(f_x, f_y)$ to $(p_x, p_y)$ w/ BFS\label{dc2g:backtrack_to_frontier}\\
	}
	\Else{
		$\mathbf{u}_{t:\infty} \leftarrow$ Shortest path to goal via BFS~\label{dc2g:backtrack_to_goal}\\
	}
	\caption{DC2G (Deep Cost-to-Go) Planner}
	\label{alg:DC2G}
\end{algorithm} 

The planning algorithm, called Deep Cost-to-Go (DC2G), is described in~\cref{alg:DC2G}.
Given the current partial semantic map, $\mathcal S$, the subset of observed cells that are also traversable (road/driveway) is $\mathcal S_{tr}$~(\cref{dc2g:traversable}).
The subset of cells in $\mathcal S_{tr}$ that are also reachable, meaning a path exists from the current position, through observed, traversable cells, is $\mathcal S_r$~(\cref{dc2g:reachable}).

The planner opts to explore if the goal cell is not yet reachable~(\cref{dc2g:if_goal_reachable}).
The current partial semantic map, $\mathcal S$, is scaled and passed into the image-to-image translator, which produces a 256x256 RGB image of the estimated cost-to-go, $\mathcal{\hat{S}}_{c2g}$~(\cref{dc2g:query}).
The raw output from the U-Net is converted to HSV-space and pixels with high value (not grayscale $\Rightarrow$ estimated not traversable) are filtered out.
The remaining grayscale image is resized to match the gridmap's dimensions with a nearest-neighbor interpolation.
The value of every grid cell in the map is assigned to be the saturation of that pixel in the translated image (high saturation $\Rightarrow$ ``whiter'' in grayscale $\Rightarrow$ closer to goal).

To enforce exploration, the only cells considered as possible subgoals are ones which will allow sight beyond frontier cells, $\mathcal F_r^e$, (reachable, traversable, and frontier-expanding), based on the known sensing range and FOV.
The cell in $\mathcal F_r^e$ with highest estimated value is selected as the subgoal~(\cref{dc2g:min_cost}).
Since the graph of reachable cells was already searched, the shortest path from the selected frontier cell to the current cell is available by backtracking through the search tree~(\cref{dc2g:backtrack_to_frontier}).
This backtracking procedure produces the list of actions, $\mathbf{u}_{t:\infty}$ that leads to the selected frontier cell.
The first action, $\mathbf{u}_{t}$ is implemented and the agent takes a step, updates its map with new sensor data, and the sense-plan-act cycle repeats.
If the goal is deemed reachable, exploration halts and the shortest path to the goal is implemented~(\cref{dc2g:backtrack_to_goal}).
However, if the goal has been observed, but a traversable path to it does not yet exist in the map, exploration continues in the hope of finding a path to the goal.

A key benefit of the DC2G planning algorithm is that it can be used alongside a local collision avoidance algorithm, which is critical for domains with dynamic obstacles (e.g., pedestrians on sidewalks).
This flexibility contrasts end-to-end learning approaches where collision avoidance either must be part of the objective during learning (further increasing training complexity) or must be somehow combined with the policy in a way that differs from the trained policy.

Moreover, a benefit of map creation during exploration is the possibility for the algorithm to confidently ``give up'' if it fully explored the environment without finding the goal.
This idea would be difficult to implement as part of a non-mapping exploration algorithm, and could address ill-posed exploration problems that exist in real environments (e.g., if no path exists to destination).

\section{Results} \label{sec:results}

\begin{figure}[t]
	\centering
	\begin{subfigure}{0.49\columnwidth}
		\centering
		\includegraphics [trim=80 80 60 60, clip, width=0.9\linewidth, angle = 0]{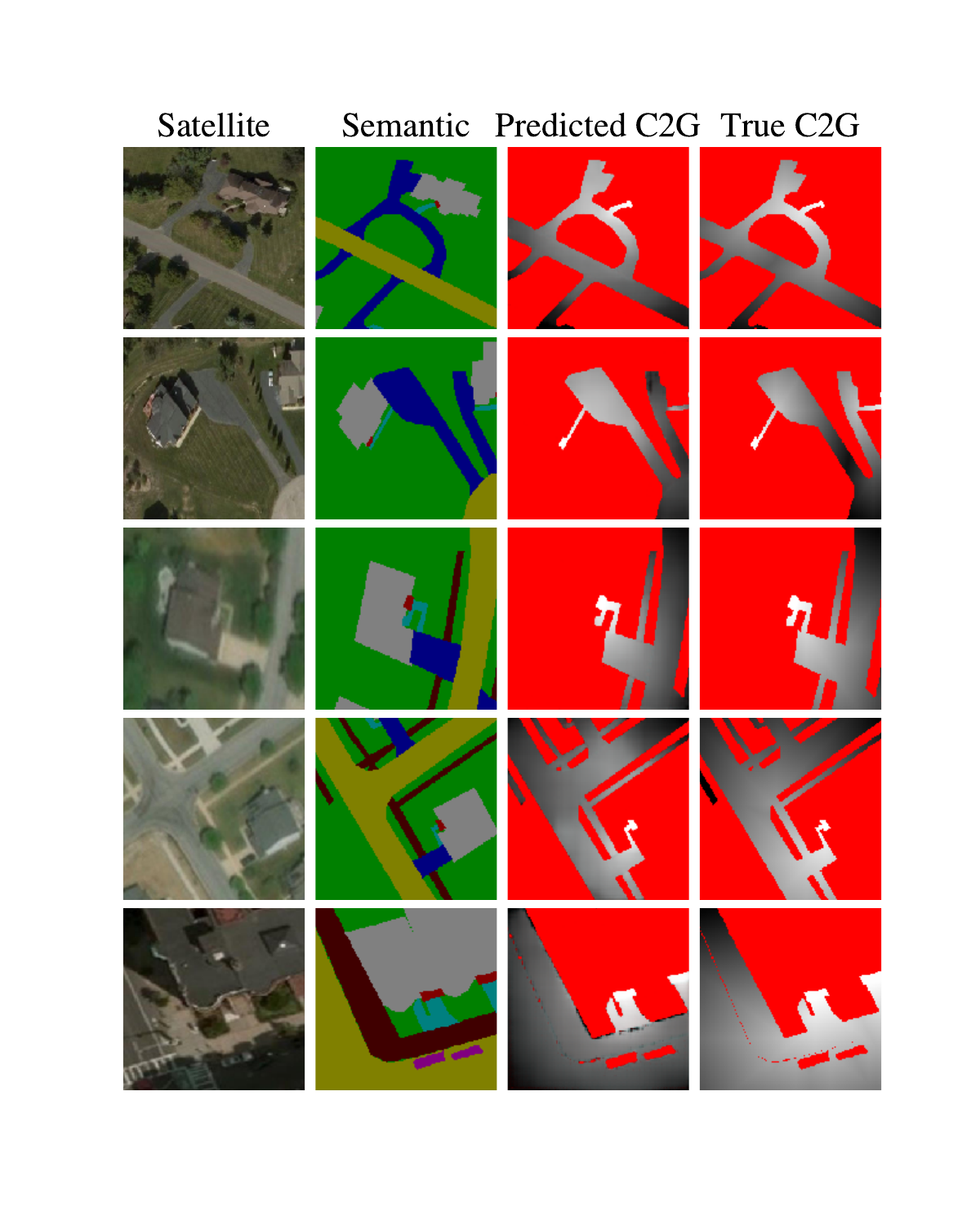}
		\caption{Full Semantic Maps}
		\label{fig:map_result_panels_full} 
	\end{subfigure}
	\centering
	\begin{subfigure}{0.49\columnwidth}
		\centering
		\includegraphics [trim=80 80 60 60, clip, width=0.9\linewidth, angle = 0]{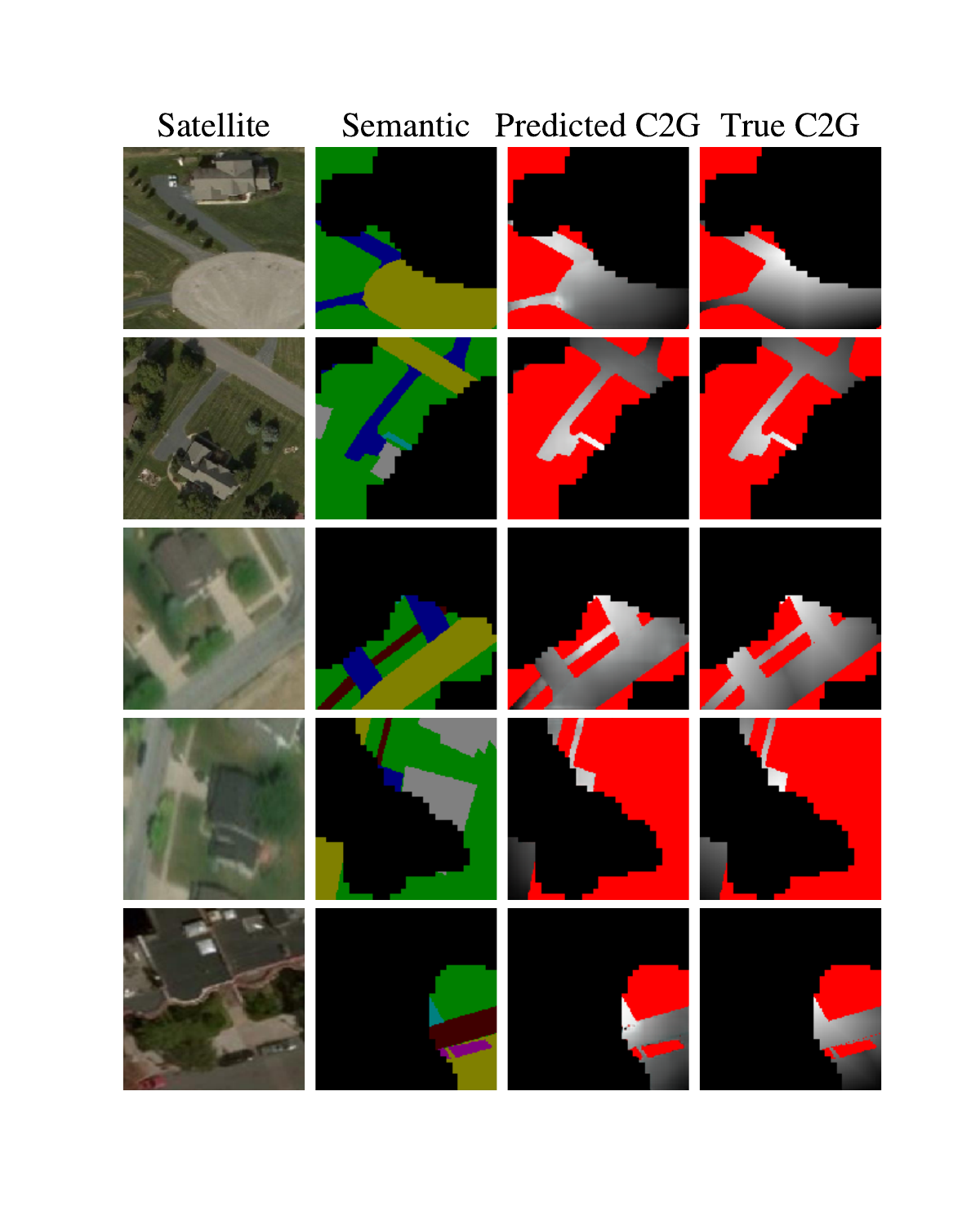}
		\caption{Partial Semantic Maps}
		\label{fig:map_result_panels_partial} 
	\end{subfigure}
	\caption{Qualitative Assessment. Network's predicted cost-to-go on 10 previously unseen semantic maps strongly resembles ground truth. Predictions correctly assign red to untraversable regions, black to unobserved regions, and grayscale with intensity corresponding to distance from goal. Terrain layouts in the top four rows (suburban houses) are more similar to the training set than the last row (urban apartments). Accordingly, network performance is best in the top four rows, but assigns too dark a value in sidewalk/road (brown/yellow) regions of bottom left map, though the traversability is still correct. The predictions in \cref{fig:map_result_panels_partial} enable planning without knowledge of the goal's location.}
	\label{fig:gan_qualitative}
\end{figure}

\begin{figure}[t]
	\centering
	\includegraphics [trim=0 0 0 0, clip, width=\linewidth, angle = 0]{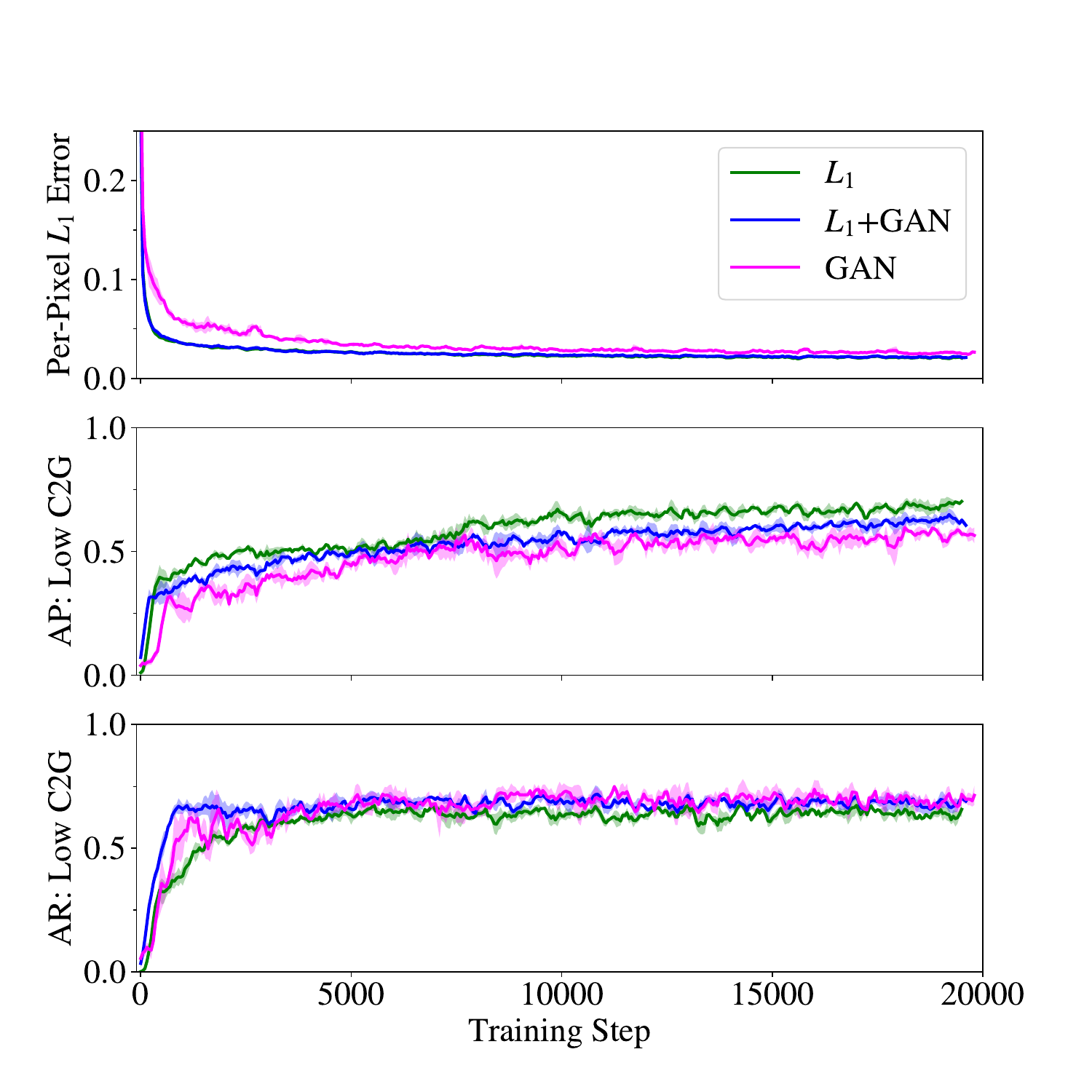}
	\caption{Comparing network loss functions. The performance on the validation set throughout training is measured with the standard $L_1$ loss, and a planning-specific metric of identifying regions of low cost-to-go. The different training loss functions yield similar performance, with slightly better precision/worse recall with $L_1$ loss (green), and slightly worse $L_1$ error with pure GAN loss (magenta). 3 training episodes per loss function are individually smoothed by a moving average filter, mean curves shown with $\pm1\sigma$ shading. These plots show that the choice of loss function has minimal impact on this work's dataset of low frequency images.}
	\label{fig:network_losses}
\end{figure}
\begin{figure}[t]
	\centering
	\includegraphics [trim=0 0 0 0, clip, width=\linewidth, angle = 0]{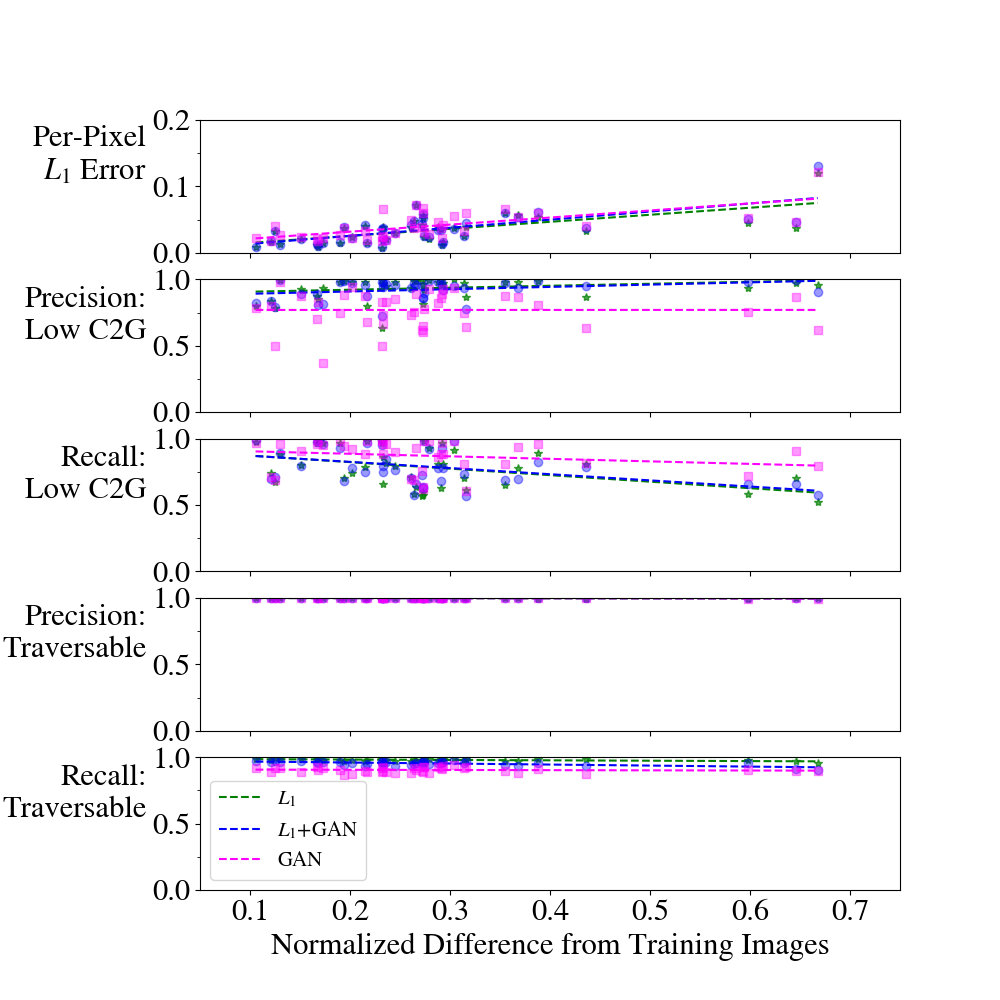}
	\caption{Cost-to-Go predictions across test set. Each marker shows the network performance on a full test house image, placed horizontally by the house's similarity to the training houses. Dashed lines show linear best-fit per network. All three networks have similar per-pixel $L_1$ loss (top row). But, for identifying regions with low cost-to-go (\nth{2}, \nth{3} rows), GAN (magenta) has worse precision/better recall than networks trained with $L_1$ loss (blue, green). All networks achieve high performance on estimating traversability (\nth{4}, \nth{5} rows). This result could inform application-specific training dataset creation, by ensuring desired test images occur near the left side of the plot.}
	\label{fig:gan_quantitative}
\end{figure}

\subsection{Model Evaluation: Image-to-Image Translation}

Training the cost-to-go estimator took about 1 hour on a GTX 1060, for 3 epochs (20,000 steps) with batch size of 1.
Generated outputs resemble the analytical cost-to-gos within a few minutes of training, but images look sharper/more accurate as training time continues.
This notion is quantified in~\cref{fig:network_losses}, where generated images are compared pixel-to-pixel to the true images in the validation set throughout the training process.

\subsubsection{Loss Functions}
Networks trained with three different loss functions are compared on three metrics in~\cref{fig:network_losses}.
The network trained with $L_1$ loss (green) has the best precision/worst recall on identification of regions of low cost-to-go (described below).
The GAN acheives almost the same $L_1$ loss, albeit slightly slower, than the networks trained explicity to optimize for $L_1$ loss.
Overall, the three networks perform similarly, suggesting the choice of loss function has minimal impact on this work's dataset of low frequency images.

\subsubsection{Qualitative Assessment}
\cref{fig:gan_qualitative} shows the generator's output on 5 full and 5 partial semantic maps from worlds and observation masks not seen during training.
The similarity to the ground truth values qualitatively suggests that the generative network successfully learned the ideas of traversability and contextual clues for goal proximity.
Some undesirable features exist, like missed assignment of light regions in the bottom row of~\cref{fig:gan_qualitative}, which is not surprising because in the training houses (suburban), roads and sidewalks are usually far from the front door, but this urban house has quite different topology.

\subsubsection{Quantitative Assessment}
In general, quantifying the performance of image-to-image translators is difficult~\cite{salimans2016improved}.
Common approaches use humans or pass the output into a segmentation algorithm that was trained on real images~\cite{salimans2016improved}; but, the first approach is not scalable and the second does not apply here.

Unique to this paper's domain\footnote{As opposed to popular image translation tasks, like sketch-to-image, or day-to-night, where there is a distribution of acceptable outputs.}, for fully-observed maps, ground truth and generated images can be directly compared, since there is a single solution to each pixel's target intensity.
\cref{fig:gan_quantitative} quantifies the trained network's performance in 3 ways, on all 42 test images, plotted against the test image's similarity to the training images (explained below).
First, the average per-pixel $L_1$ error between the predicted and true cost-to-go images is below 0.15 for all images.
To evaluate on a metric more related to planning, the predictions and targets are split into two categories: pixels that are deemed traversable (HSV-space: $S<0.3$) or not.
This binary classification is just a means of quantifying how well the network learned a relevant sub-skill of cost-to-go prediction; the network was not trained on this objective.
Still, the results show precision above 0.95 and recall above 0.8 for all test images.
A third metric assigns pixels to the class ``low cost-to-go'' if sufficiently bright (HSV-space: $V>0.9$ $\land$ $S<0.3$).
Precision and recall on this binary classification task indicate how well the network finds regions that are close to the destination.
The networks perform well on precision (above 0.75 on average), but not as well on recall, meaning the networks missed some areas of low cost-to-go, but did not spuriously assign many low cost-to-go pixels.
This is particularly evident in the bottom row of~\cref{fig:map_result_panels_full}, where the path leading to the door is correctly assigned light pixels, but the sidewalk and road are incorrectly assigned darker values.

To measure generalizability in cost-to-go estimates, the test images are each assigned a similarity score $\in [0,1]$ to the closest training image.
The score is based on Bag of Words with custom features\footnote{The commonly-used SIFT, SURF, and ORB algorithms did not provide many features in this work's low-texture semantic maps}, computed by breaking the training maps into grids, computing a color histogram per grid cell, then clustering to produce a vocabulary of the top 20 features.
Each image is then represented as a normalized histogram of words, and the minimum $L_1$ distance to a training image is assigned that test image's score.
This metric captures whether a test image has many colors/small regions in common with a training image.

The network performance on $L_1$ error, and recall of low cost-to-go regions, declines as images differ from the training set, as expected.
However, traversabilty and precision of low cost-to-go regions were not sensitive to this parameter.

Without access to the true distribution of cost-to-go maps conditioned on a partial semantic map, this work evaluates the network's performance on partial maps indirectly, by measuring the \textit{planner's} performance, which would suffer if given poor cost-to-go estimates.

\subsection{Low-Fidelity Planner Evaluation: Gridworld Simulation}

The low-fidelity simulator uses a 50$\times$50-cell gridworld~\cite{gym_minigrid} to approximate a real robotic vehicle operating in a delivery context.
Each grid cell is assigned a static class (house, driveway, etc.); this terrain information is useful both as context to find the destination, but also to enforce the real-world constraint that robots should not drive across houses' front lawns.
The agent can read the type of any grid cell within its sensor FOV (to approximate a common RGB-D sensor: $90^{\circ}$ horizontal, 8-cell radial range).
To approximate a SLAM system, the agent remembers all cells it has seen since the beginning of the episode.
At each step, the agent sends an observation to the planner containing an image of the agent's semantic map knowledge, and the agent's position and heading.
The planner selects one of three actions: go forward, or turn $\pm 90^{\circ}$.

\begin{figure}[t]
	\centering
	\includegraphics [trim=0 0 0 0, clip, width=\linewidth, angle = 0]{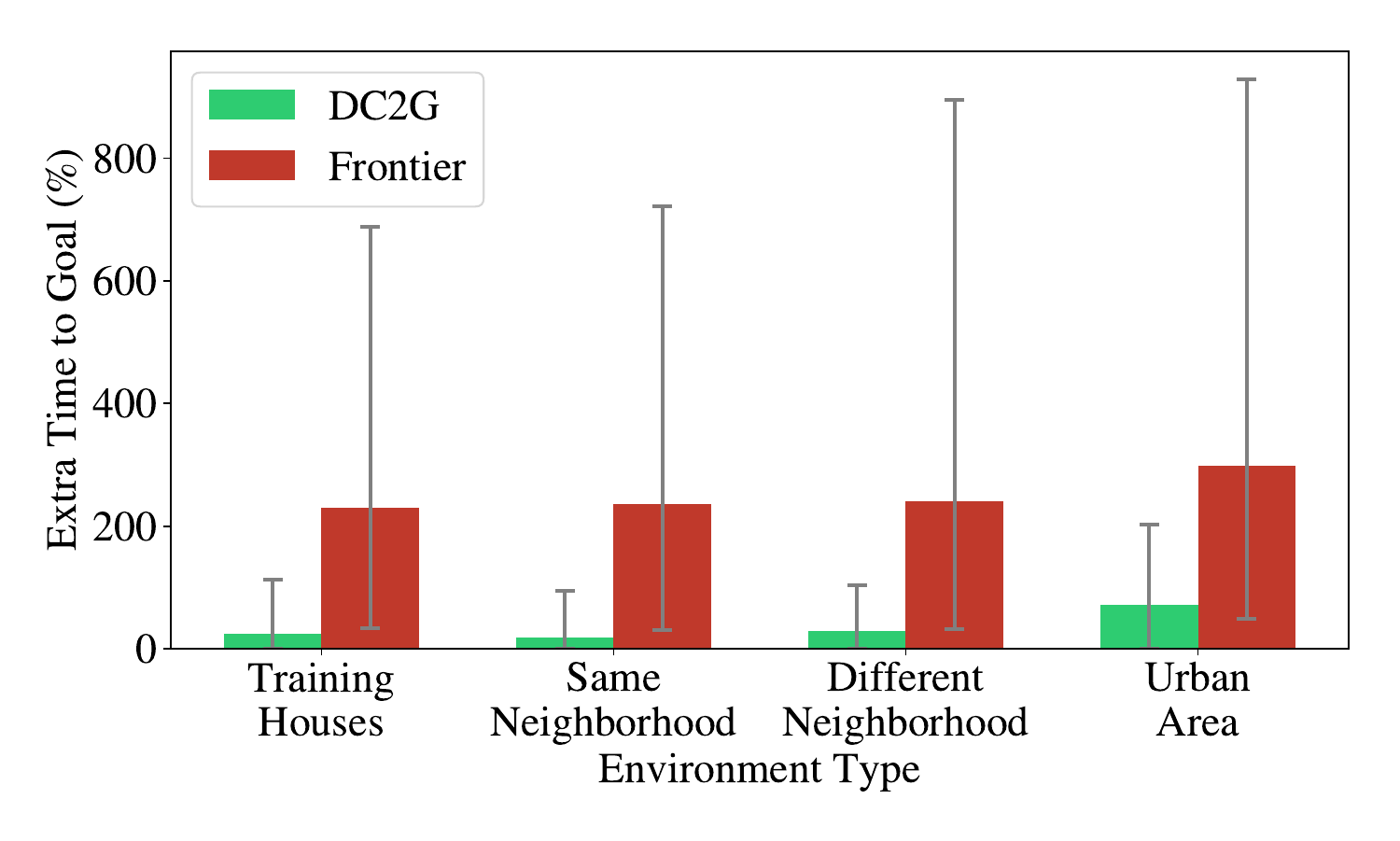}
	\caption{Planner performance across neighborhoods. DC2G reaches the goal faster than Frontier~\cite{yamauchi1998frontier} by prioritizing frontier points with a learned cost-to-go prediction. Performance measured by \% steps beyond optimal path given a complete prior map. The simulation environments come from real house layouts from Bing Maps, grouped by the four neighborhoods in the dataset, showing DC2G plans generalize beyond houses seen in training.}
	\label{fig:planner_quantitative}
\end{figure}

Each gridworld is created by loading a semantic map of a real house from Bing Maps, and houses are categorized by the 4 test neighborhoods.
A random house and starting point on the road are selected 100 times for each of the 4 neighborhoods.
The three planning algorithms compared are DC2G, Frontier~\cite{yamauchi1998frontier} which always plans to the nearest frontier cell (pure exploration), and an oracle with a prior map.
The optimal path is computed by an oracle with full knowledge of the map ahead of time; DC2G and Frontier performance is therefore presented in~\cref{fig:planner_quantitative} as percent of \textit{extra} time to the goal beyond the oracle's path, $\% t^e_{goal} = \frac{t^{alg}_{goal} - t^{oracle}_{goal}}{t^{oracle}_{goal}} \geq 0$.

\cref{fig:planner_quantitative} is grouped by neighborhood to demonstrate that DC2G improved the plans across many house types, beyond the ones it was trained on.
In the left category, the agent has seen these houses in training, but with different observation masks applied to the full map.
In the right three categories, both the houses and masks are new to the network.
Although grouping by neighborhood is not quantitative, the test-to-train image distance metric (above) was not correlated with planner performance, suggesting other factors affect plan lengths, such as topological differences in layout that were not quantified in this work.

Across the test set of 42 houses, DC2G reaches the goal within 63\% of optimal, and 189\% faster than Frontier on average.

\subsection{Planner Scenario}

\begin{figure*}[t]
	\centering
	\includegraphics [trim=0 260 100 0, clip, width=\linewidth, angle = 0, page=3]{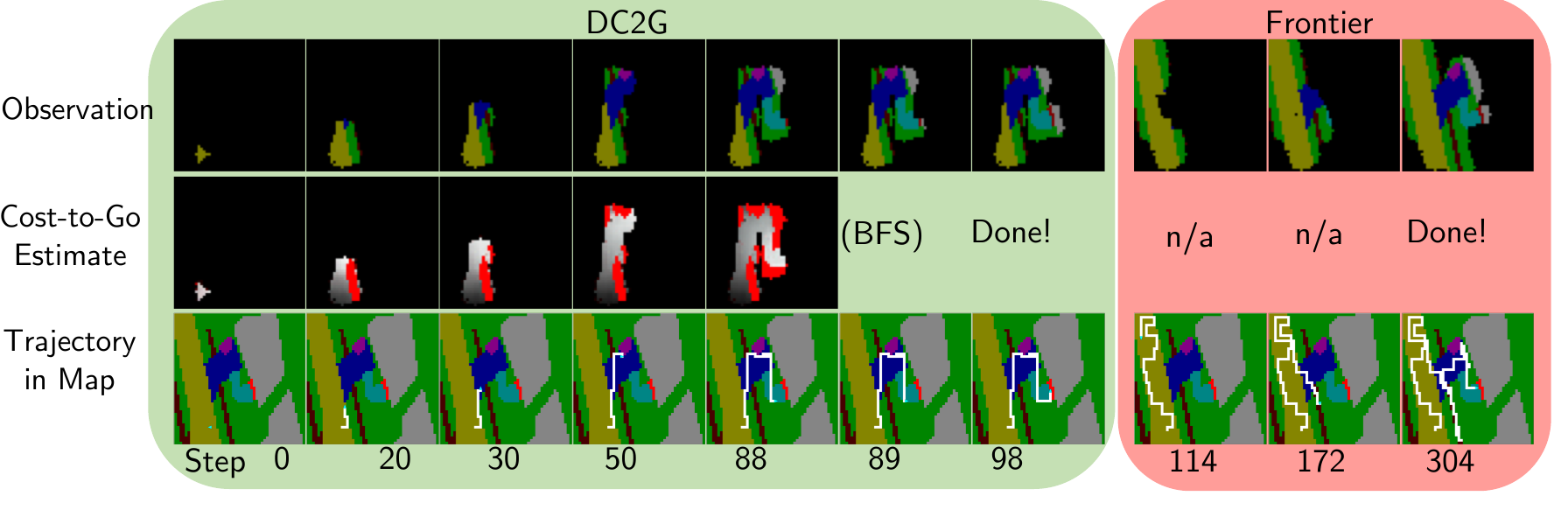}
	\caption{Sample gridworld scenario. The top row is the agent's observed semantic map (network input); the middle is its current estimate of the cost-to-go (network output); the bottom is the trajectory so far. The DC2G agent reaches the goal much faster (98 vs. 304 steps) by using learned context. At DC2G (green panel) step 20, the driveway appears in the semantic map, and the estimated cost-to-go correctly directs the search in that direction, then later up the walkway (light blue) to the door (red). Frontier (pink panel) is unaware of typical house layouts, and unnecessarily explores the road and sidewalk regions thoroughly before eventually reaching the goal.}
	\label{fig:result_trajectory}
\end{figure*}

A particular trial is shown in~\cref{fig:result_trajectory} to give insight into the performance improvement from DC2G.
Both algorithms start in the same position in a world that was not seen during training.
The top row shows the partial semantic map at that timestep (observation), the middle row is the generator's cost-to-go estimate, and the bottom shows the agent's trajectory in the whole, unobservable map.
Both algorithms begin with little context since most of the map is unobserved.
At step 20, DC2G (green box, left) has found the intersection between road (yellow) and driveway (blue): this context causes it to turn up the driveway.
By step 88, DC2G has observed the goal cell, so it simply plans the shortest path to it with BFS, finishing in 98 steps.
Conversely, Frontier (pink box, right) takes much longer (304 steps) to reach the goal, because it does not consider terrain context, and wastes many steps exploring road/sidewalk areas that are unlikely to contain a house's front door.

\subsection{Unreal Simulation \& Mapping}

\begin{figure}[t]
	\centering
	\begin{subfigure}[t]{0.8\linewidth}
		\centering
		\includegraphics [trim=0 0 0 0, clip, width=\columnwidth, angle = 0]{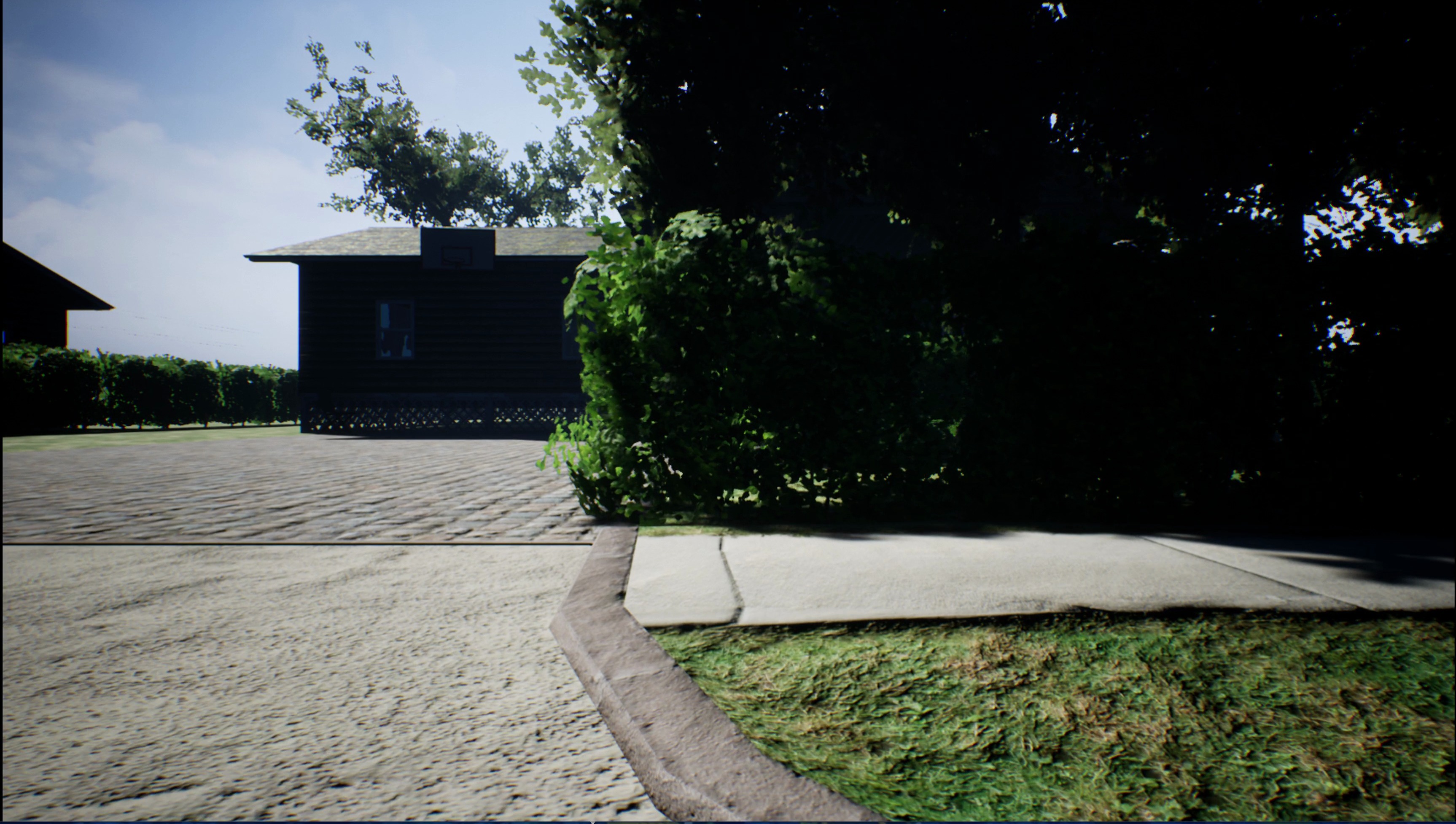}
		\caption{Unreal Engine Rendering of a Last-Mile Delivery}
		\label{fig:unreal_rendering}
		\vspace{0.1in}
	\end{subfigure}
	\begin{subfigure}[t]{0.32\columnwidth}
		\centering
		\includegraphics [trim=0 0 0 0, clip, width=\columnwidth, angle = 270]{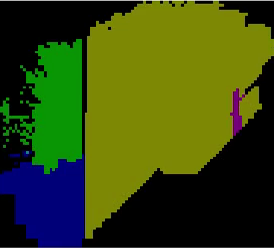}
		\caption{Semantic Map}
		\label{fig:unreal_map}
	\end{subfigure}
	\begin{subfigure}[t]{0.32\columnwidth}
		\centering
		\includegraphics [trim=0 0 0 0, clip, width=\columnwidth, angle = 270]{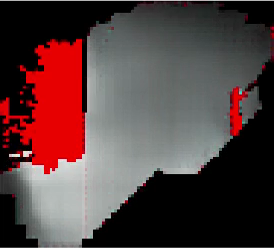}
		\caption{Predicted Cost-to-Go}
		\label{fig:unreal_c2g}
	\end{subfigure}
	\begin{subfigure}[t]{0.32\columnwidth}
		\centering
		\includegraphics [trim=0 0 0 0, clip, width=\columnwidth, angle = 270]{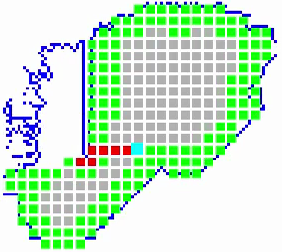}
		\caption{Planned Path}
		\label{fig:unreal_path}
	\end{subfigure}
	\caption{Unreal Simulation. Using a network trained only on aerial images, the planner guides the robot toward the front door of a house not seen before, with a forward-facing camera. The planned path (d) starts from the robot's current position (cyan) along grid cells (red) to the frontier-expanding point (green) with lowest estimated cost-to-go (c), up the driveway.}
	\label{fig:unreal}
\end{figure}

The gridworld is sufficiently complex to demonstrate the fundamental limitations of a baseline algorithm, and also allows quantifiable analysis over many test houses.
However, to navigate in a real delivery environment, a robot with a forward-facing camera needs additional capabilities not apparent in a gridworld.
Therefore, this work demonstrates the algorithm in a high-fidelity simulation of a neighborhood, using AirSim~\cite{shah2018airsim} on Unreal Engine.
The simulator returns camera images in RGB, depth, and semantic mask formats, and the mapping software described in~\cref{sec:approach} generates the top-down semantic map.

A rendering of one house in the neighborhood is shown in~\cref{fig:unreal_rendering}.
This view highlights the difficulty of the problem, since the goal (front door) is occluded by the trees, and the map in~\cref{fig:unreal_map} is rather sparse.
The semantic maps from the mapping software are much noisier than the perfect maps the network was trained with.
Still, the predicted cost-to-go in~\cref{fig:unreal_c2g} is lightest in the driveway area, causing the planned path in~\cref{fig:unreal_path} (red) from the agent's starting position (cyan) to move up the driveway, instead of exploring more of the road.
The video shows trajectories of two delivery simulations: \url{https://youtu.be/yVlnbqEFct0}.

\subsection{Discussion \& Future Work}
It is important to note that DC2G is expected to perform worse than Frontier if the test environment differs significantly from the training set, since context is task-specific.
The dataset covers several neighborhoods, but in practice, the training dataset could be expanded/tailored to application-specific preferences.
In this dataset, Frontier outperformed DC2G on $2/77$ houses.
However, even in case of Frontier outperforming DC2G, DC2G is still guaranteed to find the goal eventually (after exploring all frontier cells), unlike a system trained end-to-end that could get stuck in a local optimum.

The DC2G algorithm as described requires knowledge of the environment's dimensions; however, in practice, a maximum search area would likely already be defined relative to the robot's starting position (e.g., to ensure operation time less than battery life).
Future work will explicitly model the uncertainty in the network's cost-to-go predictions to inform a probabilistic planner, and will involve hardware experiments.
\section{Conclusion} \label{sec:conclusion}
This work presented an algorithm for learning to utilize context in a structured environment in order to inform an exploration-based planner.
The new approach, called Deep Cost-to-Go (DC2G), represents scene context in a semantic gridmap, learns to estimate which areas are beneficial to explore to quickly reach the goal, and then plans toward promising regions in the map.
The efficient training algorithm requires zero training in simulation: a context extraction model is trained on a static dataset, and the creation of the dataset is highly automated.
The algorithm outperforms pure frontier exploration by 189\% across 42 real test house layouts.
A high-fidelity simulation shows that the algorithm can be applied on a robot with a forward-facing camera to successfully complete last-mile deliveries.





\section*{Acknowledgment}
This work is supported by the Ford Motor Company.
\balance
\bibliographystyle{IEEEtran} 
\bibliography{biblio}
\end{document}